\documentclass[pdflatex,sn-basic,iicol]{sn-jnl}

\usepackage{graphicx}%
\usepackage{multirow}%
\usepackage{amsmath,amssymb,amsfonts}%
\usepackage{amsthm}%
\usepackage{mathrsfs}%
\usepackage[title]{appendix}%
\usepackage{xcolor}%
\usepackage{textcomp}%
\usepackage{manyfoot}%
\usepackage{booktabs}%
\usepackage{algorithm}%
\usepackage{algorithmicx}%
\usepackage{algpseudocode}%
\usepackage{listings}%
\usepackage{caption}%
\usepackage{lineno}%

\theoremstyle{thmstyleone}%
\theoremstyle{thmstyletwo}%

\theoremstyle{thmstylethree}%

\begin{document}

\title[Article Title]{
\centering

Globally Localizing Lunar Rover in Pixels via Graph Alignment}


\author[1,2]{\fnm{Mao} \sur{Chen}}

\author*[1,2]{\fnm{Xu} \sur{Yang}}\email{xu.yang@ia.ac.cn}

\author*[3]{\fnm{Chuankai} \sur{Liu}}\email{ckliu2005@126.com}

\author[1,2]{\fnm{Xiangkai} \sur{Zhang}}

\author[3]{\fnm{Xiaoxue} \sur{Wang}}

\author[4]{\fnm{Zheng} \sur{Bo}}

\author[3]{\fnm{Zuoyu} \sur{Zhang}}

\author[1,2]{\fnm{Zhiyong} \sur{Liu}}

\affil[1]{\orgdiv{The State Key Laboratory of Multimodal Artificial Intelligence Systems}, \orgname{Institute of Automation, Chinese Academy of Sciences}, \orgaddress{ \city{Beijing}, \country{China}}}

\affil[2]{\orgdiv{School of Artificial Intelligence}, \orgname{University of Chinese Academy of Sciences}, \orgaddress{\city{Beijing}, \country{China}}}

\affil[3]{\orgname{Beijing Aerospace Control Center}, \orgaddress{\city{Beijing}, \country{China}}}

\affil[4]{\orgname{Technology and Engineering Center for Space Utilization, Chinese Academy of Sciences}, \orgaddress{ \city{Beijing}, \country{China}}}


\abstract{
Precise rover localization is a prerequisite for autonomous lunar exploration, yet the absence of Global Navigation Satellite System (GNSS) and the cumulative drift of local localization methods severely constrain long-range missions. 
Cross-view localization provides a promising drift-free global solution by matching rover-view and satellite-view imagery. 
However, the lunar environment imposes unique challenges on correspondence alignment, including inter-entity entanglement, inter-viewpoint divergence, and simulation-to-real domain shift.
We address these challenges by proposing \textbf{W}arped \textbf{A}lignment of \textbf{R}eprojected \textbf{G}raphs (WARG), a framework that leverages unified graph learning and reprojected graph matching to achieve robust cross-view alignment.
Pretrained on the synthetic dataset LuSNAR with an average test error of 0.32m, our model demonstrates robust zero-shot generalization to the synthetic lunar south pole region with an error of 3.63 m.
Crucially, when validated on real-world data from the YuTu-2 rover, our method achieves a localization error of 1.68 m within a 100 m $\times$ 100 m search area, a precision close to one pixel in low-resolution satellite imagery (1.40 m/pixel).
Beyond accuracy, WARG is computationally efficient, containing only 1.56M parameters (16.12\% of previous lightweight models) and operating at 5.49 Hz on an NVIDIA RTX A6000 GPU, approaching GNSS-level frequency.
Finally, we observe that the model naturally develops low-level spatial awareness, such as semantic segmentation and structural reasoning, through cross-view localization learning, highlighting its potential as a promising paradigm for spatial intelligence with minimal annotation cost.
The source codes are available at \href{https://github.com/maochen-casia/warg}{https://github.com/maochen-casia/warg}.
}

\keywords{Rover localization, cross-view localization, graph alignment}



\maketitle


\section{Main}\label{sec:introduction}

Lunar exploration represents a pivotal milestone in humanity’s pursuit to extend its reach beyond Earth, offering unparalleled opportunities to advance science, technology, and sustainable space operations~\citep{challenges,cesar}. 
Motivated by both scientific curiosity and strategic vision, nations around the world have launched ambitious lunar programmes. 
China’s Chang’e missions have demonstrated remarkable achievements, from the first far-side landing to successful sample return~\citep{chang-e-6-1,chang-e-6-2}, while NASA’s Artemis programme aims to establish a long-term lunar base as a stepping stone toward Mars~\citep{artemis-1,artemis-2}. 
Similarly, ESA’s Heracles~\citep{heracles} and ISRO’s Chandrayaan~\citep{chandrayaan} missions highlight a growing international commitment to sustained lunar research and in-situ resource utilisation.
At the heart of these endeavours lies a crucial operational challenge: accurately localizing lunar rovers, the primary agents of surface exploration. 
Reliable localization underpins key activities such as autonomous navigation across unstructured terrains and precise geological sampling. 
However, achieving this remains difficult in the Moon’s feature-sparse, GNSS-denied environment~\citep{crater,lunarnav}. 
Developing localization techniques that can operate reliably under these extreme conditions is therefore essential to ensure both the safety and the scientific return of future missions.

\begin{figure}[!t]
    \centering
    \includegraphics[width=0.9\linewidth]{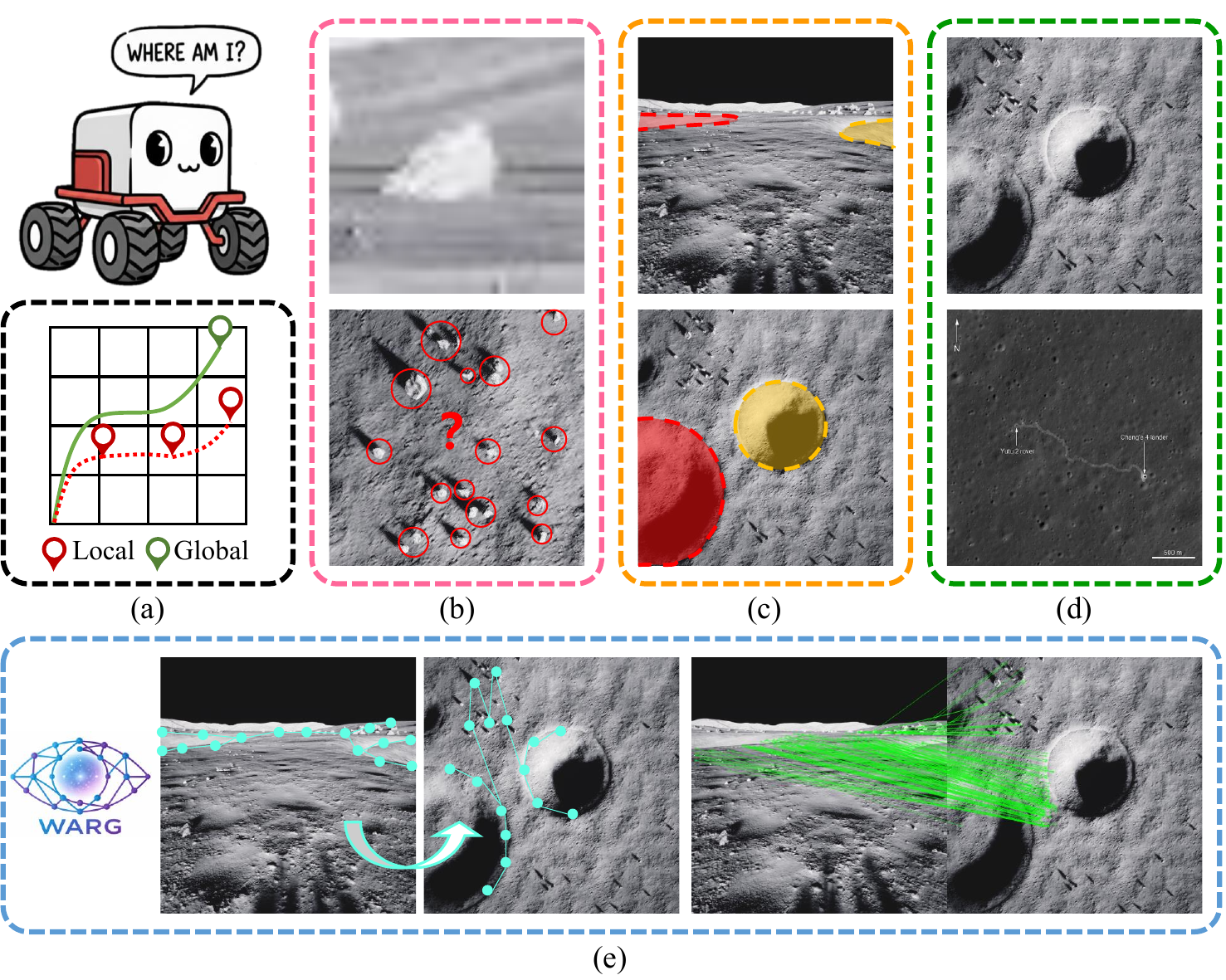}
    \caption{Overview of lunar localization and challenges. (a) Local vs. global localization. (b) Inter-entity entanglement. (c) Intra-entity divergence. (d) Sim-to-real gap. (e) WARG achieves accurate warped alignment based on graph reprojection.}
    \label{fig:intro}
\end{figure}

Unlike on Earth, where the GNSS provides reliable global and absolute localization, lunar rovers operate in a completely GNSS-denied environment~\citep{gnss}, relying solely on onboard sensors. 
Previous lunar missions primarily adopted local localization methods, which estimate the rover’s motion with respect to a local reference frame, such as its previous position, as shown in Fig.~\ref{fig:intro}(a).
Typical approaches include inertial navigation and wheel odometry~\citep{yutu-2,chandrayaan-3}, which infer motion from accelerometer, gyroscope, and wheel encoder measurements.
However, they are highly sensitive to sensor noise, bias drift, and wheel slippage, leading to cumulative errors in estimated position~\citep{inertial}.
To improve robustness, stereo-based localization has been introduced, exploiting stereo vision to reconstruct sparse 3D points and estimate the rover’s relative motion between consecutive frames through feature or point cloud matching~\citep{stereo-1,stereo-2}.
This technique improves short-term accuracy and robustness compared to purely inertial systems.
Despite achieving short-term accuracy within one meter, all these relative localization methods suffer from accumulated drift over long traverses~\citep{stereo-drift}, as even modest sensor noise or minor errors in depth estimation and image correspondence can propagate over time and lead to kilometer-level position deviations after extended operations, thereby constraining their applicability for long-term and large-scale lunar exploration~\citep{chang-e-7}.

To overcome the limitations of local localization, which also hinder navigation and autonomous driving in some GNSS-denied environments on Earth, recent studies~\citep{orienternet,ccvpe,fg2} have shifted toward cross-view localization, a paradigm that achieves global localization analogous to GNSS particularly in terrestrial settings.
When extended to lunar rovers, this paradigm may leverage cross-view information between rover and satellite imagery, rather than relying on consecutive rover frames, to establish spatial correspondences.
Specifically, by matching visual features captured from the rover’s viewpoint with their counterparts in satellite images, the rover’s absolute position within a global reference frame can be directly determined.
This strategy is inherently resistant to drift, as each estimate is independent of prior motion measurements, as illustrated in Fig.~\ref{fig:intro}(a). 
As a result, cross-view localization offers a promising foundation for achieving long-term accuracy and scalability in future lunar exploration missions.

Despite their promise, existing cross-view localization methods still struggle to achieve high accuracy when applied to lunar scenarios, often exhibiting localization errors exceeding tens of meters, which renders them impractical for real lunar deployment.
The substantial challenges stem mainly from the Moon’s repetitive terrain and the extreme discrepancy between rover and satellite viewpoints.
First, the repetitiveness of lunar terrain leads to the inter-entity entanglement, where distinct entities, such as two craters separated by 1 km, may look identical, as do countless small rocks shown in Fig.~\ref{fig:intro}(b).
Most existing methods~\citep{fg2,loc2} typically rely on entity-wise alignment, finding a specific match for a given feature in another view.
While this works well on Earth with discriminative landmarks like buildings, it fails on the Moon, where satellite imagery often presents multiple, visually ambiguous candidates for a single target.
Second, the significant viewpoint discrepancies cause intra-entity divergence, where a single entity exhibits distinct appearances across the two viewpoints, as demonstrated in Fig.~\ref{fig:intro}(c).
The extreme scale difference between rover and satellite imagery leads to geometric distortion and partial observability.
For example, a massive crater seen from satellite imagery may be only partially visible in the rover's imagery.
Furthermore, drastic illumination variations, particularly in the lunar south pole region~\citep{illumination}, significantly alter entities' visual features.
Most existing methods~\citep{orienternet,slicematch,hc-net} process the two viewpoints independently, which compels the model to prioritize viewpoint-specific details over shared, intrinsic characteristics.
Consequently, they fail to extract the scale- and illumination-invariant features necessary to mitigate the intra-entity divergence.
Third, as shown in Fig.~\ref{fig:intro}(d), the domain shift derived from the simulation-to-real gap limits practical deployment.
Due to the scarcity of real lunar data, most existing data-intensive deep learning models are trained primarily on synthetic data~\citep{lcv-1,lcv-2,lunarloc}.
However, given their architectural complexity and high parameter counts, these methods are prone to overfitting the simulation environment and often fail to generalize to real-world conditions.

In this work, we propose \textbf{W}arped \textbf{A}lignment of \textbf{R}eprojected \textbf{G}raphs (WARG) to achieve accurate cross-view rover localization as shown in Fig.~\ref{fig:intro}(e). 
First, WARG leverages the graph representation to resolve inter-entity entanglement, as the structure of multiple entities provides more discriminative features than individual entities alone.
Second, WARG employs a weight-sharing architecture to process both viewpoints symmetrically, compelling the model to disregard viewpoint-specific noise and overcome inter-viewpoint divergence.
Third, WARG’s efficient architecture and integration of geometric reprojection priors enhance its domain generalization, enabling high performance even when fine-tuned on a limited amount of real-world data.

Our findings indicate that leveraging these properties effectively addresses the aforementioned challenges and yields substantial gains in accuracy, efficiency, and generalization.
Moreover, WARG demonstrates strong robustness under various extreme conditions, including severe occlusion and blur.
Finally, we observe that through simple cross-view localization learning, our model naturally develops low-level spatial awareness, exhibiting emergent generalization to tasks that require semantic and structural reasoning.

\subsection{Contributions}
First, we propose WARG, an effective and efficient graph-based cross-view localization method specifically designed for the challenging lunar environment.
By integrating graph learning with geometric reprojection, our method achieves precise alignment between rover and satellite viewpoints, effectively addressing the unique challenges of lunar terrain and providing accurate global localization for future large-scale lunar exploration missions.

Second, WARG delivers a significant performance leap over existing methods while maintaining high computational efficiency.
In a 100 m $\times$ 100 m search region, WARG reduces the mean localization error from 20.89 m to 0.32 m in the standard lunar simulation.
It further demonstrates strong zero-shot capability, achieving 3.63 m error in the complex south pole simulation, a nearly ten-fold improvement over the 
33.09 m for previous methods.
When fine-tuned on limited real-world data from the YuTu-2, our model achieves a 1.68 m error (nearly 1 pixel at 1.40 m/px resolution), significantly outperforming the prior state-of-the-art (SOTA) of 12.89 m.
Moreover, it supports a processing rate of 5.49 high-resolution (1024 px) image pairs per second on an NVIDIA RTX A6000 GPU, approaching GNSS-level frequency.

Third, WARG exhibits exceptional robustness against environmental stressors. 
It maintains high accuracy under extreme conditions, such as severe occlusion, blur, and drastic illumination or scale discrepancies between viewpoints.
This resilience confirms its readiness for practical, large-scale lunar exploration missions.

Finally, through the simple cross-view localization task, our model naturally develops low-level spatial awareness, encompassing both semantic and structural reasoning ability essential for tasks like semantic segmentation and image correspondence.
This finding suggests that cross-view localization can serve as a promising pathway toward spatial intelligence, enabling models to acquire diverse spatial reasoning capabilities with minimal labeling effort, thereby bypassing the prohibitive annotation costs typically associated with dense supervision.


\section{Results}

In this section, we first demonstrate the effectiveness and efficiency of the proposed WARG on lunar cross-view localization and focus on its ability to overcome the fundamental challenges of the lunar environment.
We then analyze the emergent low-level spatial awareness that the model develops during this process.
To assess its viability for real-world deployment, we evaluate its operational efficiency and robustness under varying conditions.
Finally, we provide a detailed ablation study to quantify the impact of different modules and a robustness study to evaluate performance under varying conditions.

\subsection{Benchmark Performance}

\begin{figure*}[!t]
    \centering
    \includegraphics[width=0.9\linewidth]{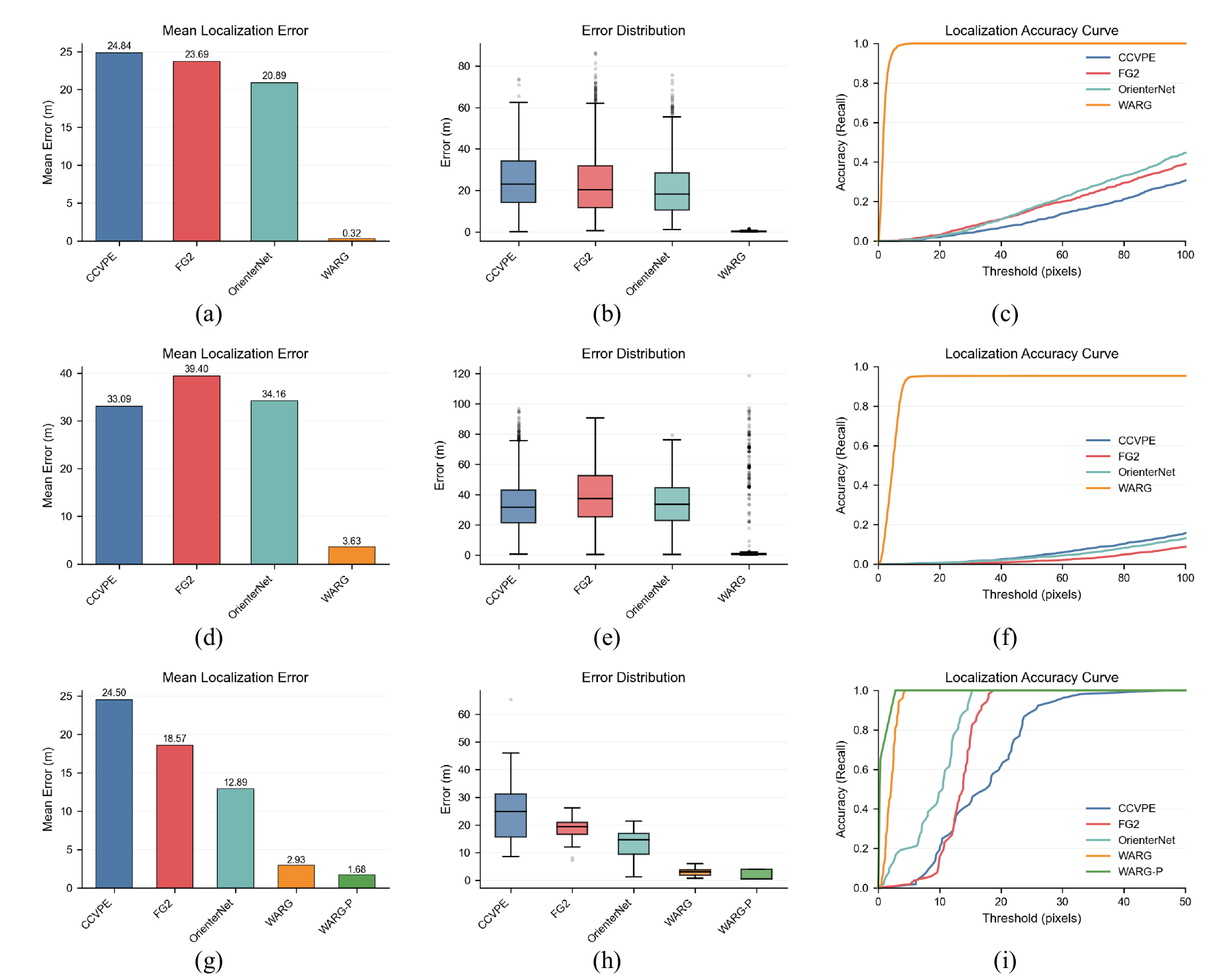}
    \caption{Localization performance comparison on three datasets in a 100 m $\times$ 100 m search region. Comparisons are shown for the LuSNAR (a–c), South (d–f), and YuTu-2 (g–i) datasets. The columns represent: (left) Mean Localization Error in meters; (middle) Error distribution box plots; and (right) Localization accuracy curves, showing recall rates across varying pixel error thresholds.}
    \label{fig:main_result}
\end{figure*}

We evaluate our framework using three complementary datasets: LuSNAR, South, and YuTu-2. 
The LuSNAR dataset provides a large-scale simulation of representative lunar terrains, comprising 10,020 training, 1,625 validation, and 1,350 test images across diverse geological scenes. 
To assess the model's generalization capabilities under challenging conditions, we utilize the synthetic South dataset, which replicates the extreme illumination of the lunar south pole and serves as a 3,000-image zero-shot benchmark. 
Finally, we validate the real-world applicability and sim-to-real transfer of our method using the YuTu-2 dataset, which consists of authentic lunar surface imagery (176 training and 52 test images) captured by the YuTu-2 rover.
Considering the sequential nature of the YuTu-2 trajectory data, we reserve the final two stops of the 11-stop trajectory for testing to prevent model interpolation.
Advanced cross-view localization methods, including OrienterNet~\citep{orienternet}, CCVPE~\citep{ccvpe}, and FG2~\citep{fg2}, are included in comparison using the same frozen backbone DINOv3~\citep{dinov3}.

Figure~\ref{fig:main_result} provides a comprehensive evaluation of WARG’s localization performance against baseline methods across three benchmarks.
In the LuSNAR dataset, which represents standard lunar environments, WARG demonstrates a substantial improvement in precision (Fig.\ref{fig:main_result}(a–c)).
Specifically, it reduces the mean localization error from the previous SOTA OrienterNet of 20.89 m to just 0.32 m (Fig.\ref{fig:main_result}(a)). 
Beyond average performance, the significant reduction in outliers shown in Fig.\ref{fig:main_result}(b) underscores the framework's superior robustness and generalization capabilities.
Notably, WARG successfully localizes over 80\% of cases within a 3-pixel threshold at a resolution of 0.16 m/pixel.
This represents an unprecedented level of sub-meter accuracy that far surpasses all existing baselines (Fig.\ref{fig:main_result}(c)).
WARG’s generalization capabilities are further underscored by its performance on the South dataset during zero-shot validation. 
While baseline methods exhibit performance degradation approaching the expected error of random localization (approximately 38.26 m), WARG maintains a high-precision mean localization error of 3.63 m (Fig.\ref{fig:main_result}(d)). 
The error distribution in Fig.\ref{fig:main_result}(e) demonstrates that most predictions are concentrated within a narrow margin.
The few observed outliers correspond to edge-case scenarios where the rover is positioned at the boundary of the satellite image, resulting in minimal spatial overlap.
Finally, we demonstrate the real-world utility of WARG using authentic imagery from the YuTu-2 mission.
As shown in Fig.\ref{fig:main_result}(g), WARG significantly outperforms existing methods, achieving a mean localization error of 2.93 m, which is a substantial improvement over the 12.89 m error recorded by the SOTA OrienterNet. 
A key advantage of our framework is its architectural versatility, which allows it to seamlessly integrate panoramic observations. 
By leveraging multiple views at a single location, the refined WARG-P model further suppresses the localization error to 1.68 m. 
At a satellite resolution of 1.40 m/pixel, this represents remarkable precision, approaching one pixel. 
Furthermore, the error distribution in Fig.\ref{fig:main_result}(h) and the accuracy curves in Fig.~\ref{fig:main_result}(i) confirm that WARG maintains high operational reliability with fewer outliers and consistently higher recall rates across all error thresholds, establishing its readiness for actual lunar deployment.
Notably, while WARG achieves sub-meter accuracy on the disjoint LuSNAR dataset, its mean error increases to 2.93 m on the YuTu-2 dataset. 
This shift is consistent with the resolution difference between the high-fidelity simulation (0.16 m/pixel) and the coarser real orbital imagery (1.40 m/pixel), indicating that WARG maintains a stable pixel-level precision (about 2 pixels) across domains.
Conversely, baseline methods achieve better performance on the YuTu-2 dataset compared to the simulated LuSNAR dataset.
This counterintuitive result highlights their sensitivity to scale changes and inability to effectively encode high-frequency details.
Consequently, these limitations prevent the baselines from leveraging the finer details provided by next-generation high-resolution maps.

\begin{figure*}[!h]
    \centering
    \includegraphics[width=0.9\linewidth]{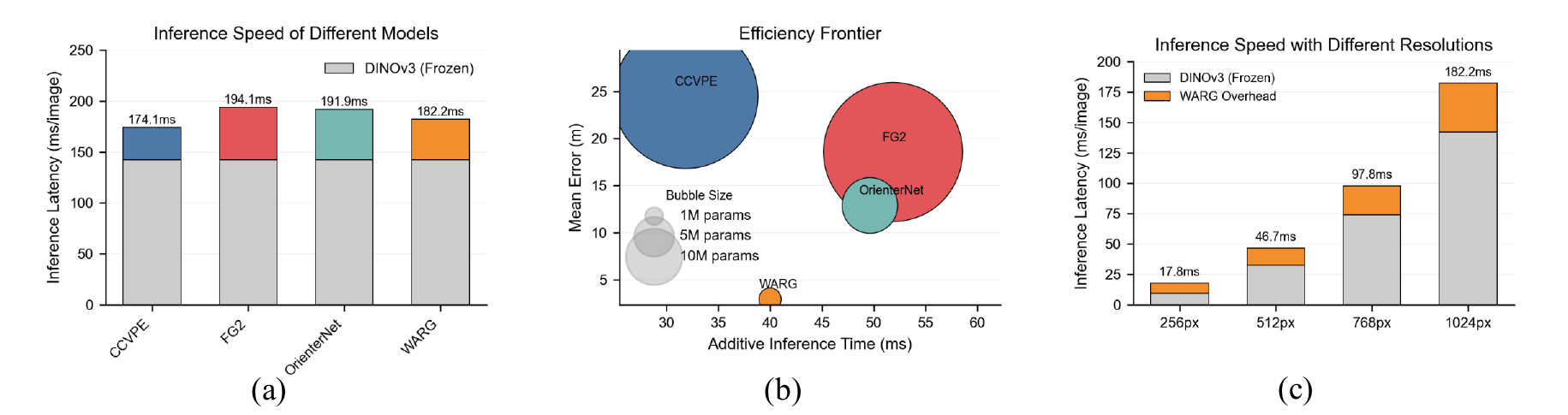}
    \caption{Computational efficiency comparison. (a) Comparison of total inference latency at 1024 $\times$ 1024 px resolution, decomposed into frozen backbone processing (DINOv3) and model-specific overhead. (b) Efficiency frontier visualizing the trade-off between mean localization error (m) and additive inference time (ms). Bubble size represents the number of trainable parameters. (c) Scalability analysis of inference speed across varying input resolutions.}
    \label{fig:efficiency}
\end{figure*}

Beyond its superior localization accuracy, WARG exhibits high computational efficiency.
As illustrated in Fig.~\ref{fig:efficiency}(a), at a standard resolution of 1024 px, WARG achieves an inference speed comparable to or faster than established baselines, with its additional overhead representing only a small fraction of the total processing time. 
This efficiency is most striking when visualized as an efficiency frontier (Fig.~\ref{fig:efficiency}(b)). While baselines typically require significantly higher parameter counts or increased latency to achieve even coarse-grained precision, WARG occupies a unique optimal position, delivering a substantial reduction in mean error while maintaining a substantially smaller parameter footprint.
This streamlined architecture ensures that the model provides high-precision intelligence without the typical weight of deep-learning models, making it highly suitable for deployment on low-power spaceborne processors.
Furthermore, the framework exhibits robust resolution scalability (Fig.~\ref{fig:efficiency}(c)).
As the input resolution increases, the additional latency introduced by WARG remains proportionally low compared to the backbone. 
By naturally balancing sub-meter precision with minimal computational costs and architectural complexity, WARG establishes a new benchmark for practical, real-time autonomous navigation in lunar environments.

\begin{figure*}[!t]
    \centering
    \includegraphics[width=0.95\linewidth]{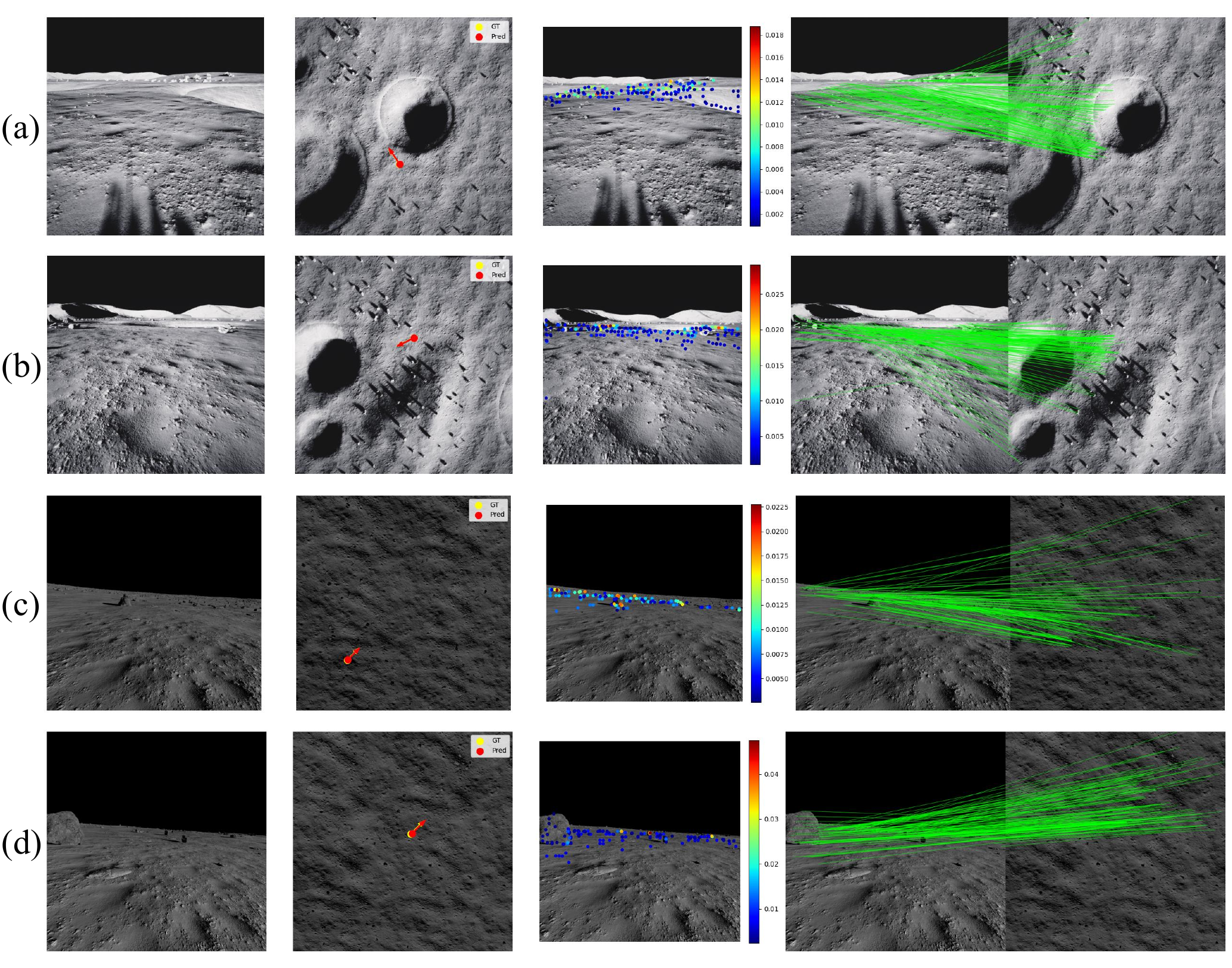}
    \caption{Visualization of WARG localization on the LuSNAR (a-b) and South (c-d) datasets. From left to right: the input rover-view imagery; the corresponding satellite-view reference map with ground-truth (yellow) and predicted (red) positions overlaid; the distribution of captured nodes in the rover view, color-coded by their saliency weights (indicating feature importance); and the final cross-view correspondences established through graph-based structural matching.}
    \label{fig:main_correspondence}
\end{figure*}

We further visualize the localization process of WARG in Fig.~\ref{fig:main_correspondence}.
Notably, a prominent challenge in lunar navigation is the significant domain gap between standard topographic regions (LuSNAR) and the extreme, shadow-dominant terrains of the lunar south pole (South).
Despite these disparate visual characteristics, WARG demonstrates remarkable zero-shot generalization, maintaining high-precision localization in both settings.
The third column of Fig.~\ref{fig:main_correspondence} reveals how the model autonomously identifies and prioritizes informative terrain features. 
Rather than processing the entire image uniformly, WARG selectively captures nodes with high saliency weights centered on stable geomorphological landmarks, such as crater rims and rocks.
This self-supervised prioritization allows the model to ignore uninformative, transient textures and focus on the underlying structural layout of the landscape.
Ultimately, the accurate correspondence matching shown in the final columns proves that WARG effectively overcomes the challenges of inter-entity entanglement and intra-entity divergence and achieves highly accurate and robust cross-view localization, ensuring reliable navigation even under the most demanding environmental conditions of the lunar surface.

\subsection{Inter-Entity Entanglement}

\begin{figure*}
    \centering
    \includegraphics[width=0.95\linewidth]{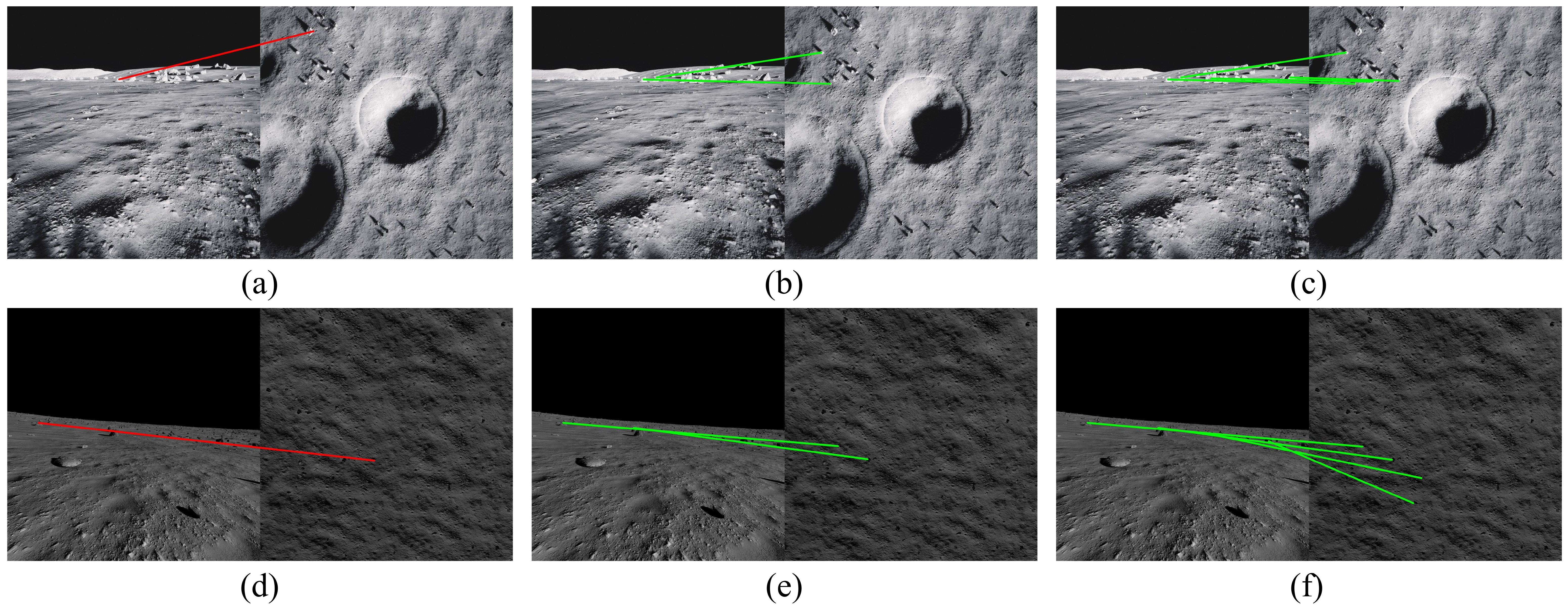}
    \caption{Graph structural constraints resolve inter-entity entanglement. Visual correspondence results between rover views (left) and satellite maps (right) on the LuSNAR (a–c) and South (d–f) datasets. The columns illustrate the impact of increasing the number of entities used for structural matching (1, 2, 4).}
    \label{fig:inter_correspondence}
\end{figure*}

\begin{figure*}
    \centering
    \includegraphics[width=0.8\linewidth]{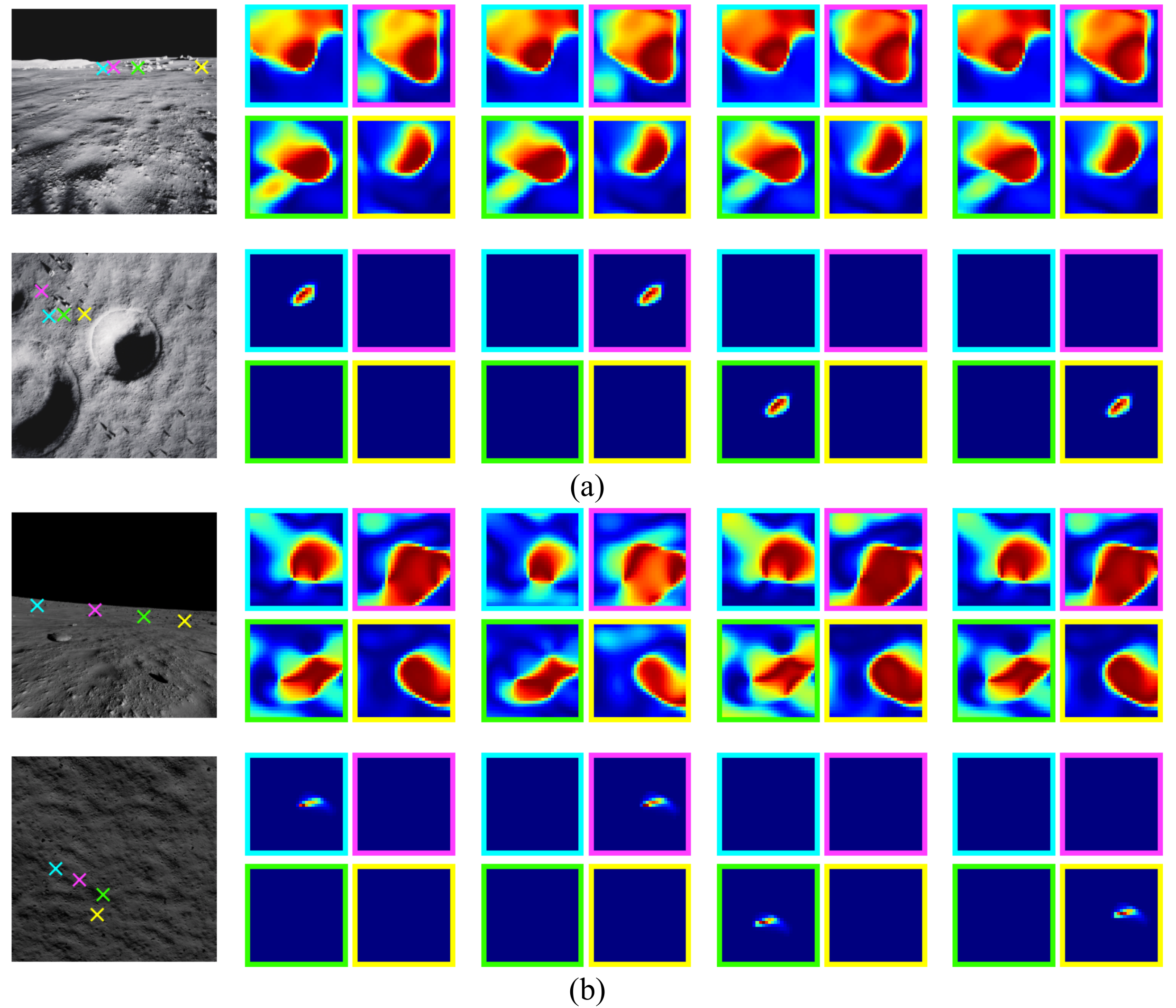}
    \caption{Graph structural constraints eliminate inter-entity ambiguity. Visualization of feature matching on the LuSNAR (a) and South datasets (b). Left: Rover view and Satellite map with four target entities marked by color-coded crosses. Right: Similarity heatmaps arranged in columns corresponding to the four entities. Each column visualizes the similarity distribution between a specific rover-view entity and the features in the satellite map. The top row displays the raw, appearance-based similarity. The bottom row shows the refined similarity after applying graph structural constraints.}
    \label{fig:inter_sim}
\end{figure*}

\begin{figure*}[!t]
    \centering
    \includegraphics[width=0.95\linewidth]{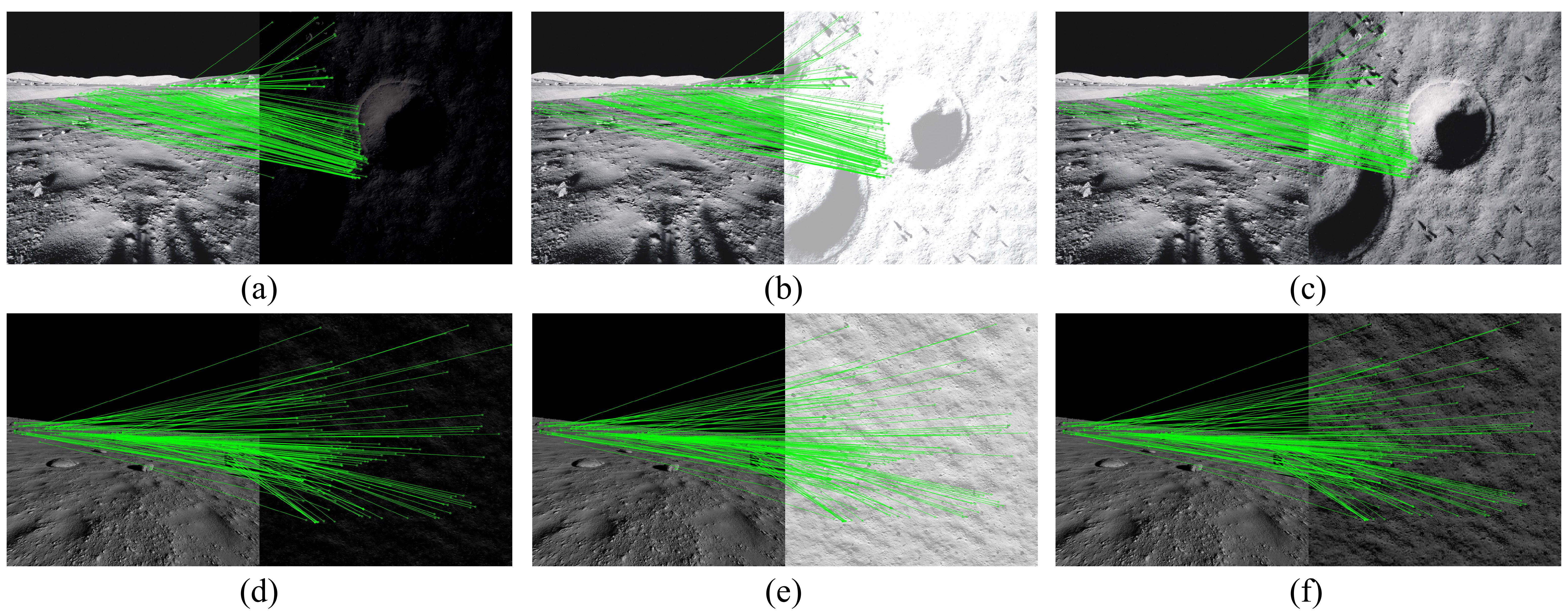}
    \caption{WARG overcomes intra-entity divergence caused by extreme illumination and scale discrepancies. Qualitative evaluation on the LuSNAR (a–c) and South (d–f) datasets. (a)(b)(d)(e), Visualization of matching performance under extreme illumination changes. (c)(f), Evaluation of matching performance using low-resolution satellite imagery (256 px), simulating the enlarged scale difference between two viewpoints.}
    \label{fig:intra_entity_correspondence}
\end{figure*}

In this subsection, we address the challenge of inter-entity entanglement, a phenomenon where distinct geological entities (e.g., craters or rocks) exhibit high visual similarity, leading to perceptual aliasing. 
We posit that compelling a model to disambiguate these semantically identical instances in isolation is an ill-posed problem that compromises generalization and encourages overfitting to specific textures. 
Instead, we propose a shift from instance-level identification to structural constellation matching.
While individual entities may lack discriminative power, their relative structure forms a unique signature that is highly robust to visual ambiguity. 
Consequently, WARG enforces structural constraints during matching to effectively resolve inter-entity entanglement.

Fig.~\ref{fig:inter_correspondence} provides a qualitative validation of our structural matching hypothesis. In the single-entity baseline (Fig.~\ref{fig:inter_correspondence}(a)(d)), the model succumbs to visual aliasing: the high visual similarity between distinct lunar rocks leads to catastrophic mismatches.
This confirms that local appearance alone is insufficient for disambiguation in repetitive terrains.
However, the introduction of structural constraints fundamentally alters the matching landscape. 
As shown in Fig.~\ref{fig:inter_correspondence}(b)(e), expanding the scope to just two entities begins to filter out false positives by imposing a pair constraint. 
By further increasing the constellation to four entities (Fig.~\ref{fig:inter_correspondence}(c)(f)), the structure becomes unique, and the relative spatial arrangement of the features forms a discriminative signature that does not exist elsewhere in the search space. 
Consequently, WARG successfully locks onto the correct correspondences across both the standard LuSNAR environment and the high-contrast South dataset, proving that structure is the key to resolving inter-entity entanglement.

To mechanistically understand how WARG resolves inter-entity entanglement, we visualize the internal similarity maps of the matching module in Fig.~\ref{fig:inter_sim}. 
The top row of heatmaps illustrates the challenge of relying solely on local appearance: the probability distributions are diffuse and multi-modal. 
For example, the specific rover entity marked in Cyan (Fig.~\ref{fig:inter_sim}(a), column 1) shows high similarity to multiple distinct regions in the satellite map, confirming that the model effectively hallucinates matches due to the repetitive nature of lunar terrain features. 
This visualizes the inter-entity entanglement, as the model cannot distinguish the true target from other similar-looking rocks based on texture alone.
In contrast, the bottom row demonstrates the decisive impact of structural constraints. 
By modeling the four entities as a graph rather than isolated points, WARG conditions the matching of each entity on the spatial structure. 
This structural context acts as a powerful filter, suppressing high-confidence false positives that are structurally inconsistent with the constellation. 
The result is a collapse of the probability distribution into a single, sharp peak for each entity, proving that structural consistency provides the discriminative power necessary to distinguish visually identical features.

\subsection{Intra-Entity Divergence}

In addition to resolving inter-entity entanglement, robust localization requires addressing intra-entity divergence caused by viewpoint discrepancies. 
A single geological entity can exhibit vastly different visual features depending on the observation angle, scale, and illumination conditions. 
WARG tackles this by adopting a weight-sharing architecture to process both rover and satellite viewpoints. 
This design explicitly encourages the extraction of viewpoint-invariant features by mapping both modalities to a common feature space. 
We explicitly investigate the impact of this strategy in overcoming the divergence inherent in cross-view data.

Fig.~\ref{fig:intra_entity_correspondence} visually confirms the effectiveness of WARG in mitigating intra-entity divergence. 
In scenarios dominated by illumination discrepancies (Fig.~\ref{fig:intra_entity_correspondence}(a-b)(d-e)), the model successfully ignores transient lighting artifacts, such as shifting shadows and high-contrast albedo changes, to lock onto the underlying geometric structure of the terrain. 
Furthermore, the results in Fig.~\ref{fig:intra_entity_correspondence}(c)(f) demonstrate remarkable scale invariance.
Even when the satellite map is downsampled from 1024 $\times$ 1024 pixels to 256 $\times$ 256 pixels, which quadruples the spatial scale (meters per pixel) and severely exacerbates the cross-view scale discrepancy, WARG maintains accurate correspondence matching.

\begin{figure*}[!t]
    \centering
    \includegraphics[width=0.95\linewidth]{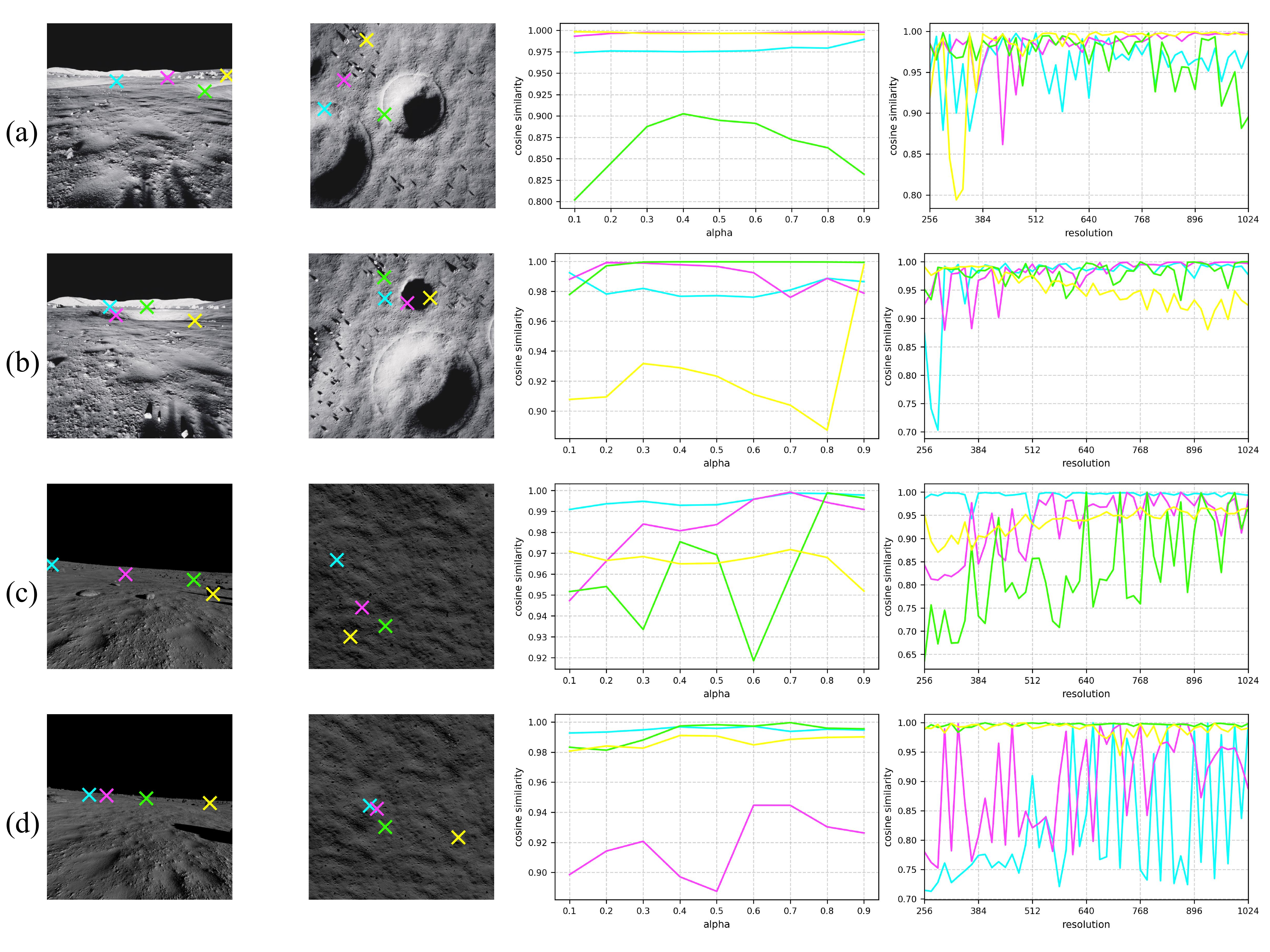}
    \caption{Quantitative evaluation of feature invariance against extreme illumination and scale perturbations. Performance is analyzed on the LuSNAR (a, b) and South (c, d) datasets. For each scene, the left panels display the rover and satellite views with four color-coded corresponding entities. The middle plots quantify the feature stability (measured by cosine similarity) as a function of illumination intensity $\alpha$, where $\alpha=0.5$ represents the original lighting conditions, $\alpha \xrightarrow{} 0$ simulates complete darkness, and $\alpha \xrightarrow{} 1$ simulates complete overexposure (whiteness). The right plots track feature similarity as the satellite image resolution is degraded from 1024 $\times$ 1024 down to 256 $\times$ 256 pixels. High cosine similarity values across these ranges indicate that WARG extracts robust, viewpoint-invariant representations despite severe environmental degradation.}
    \label{fig:intra_entity_sim}
\end{figure*}

\begin{figure*}[!t]
    \centering
    \includegraphics[width=0.95\linewidth]{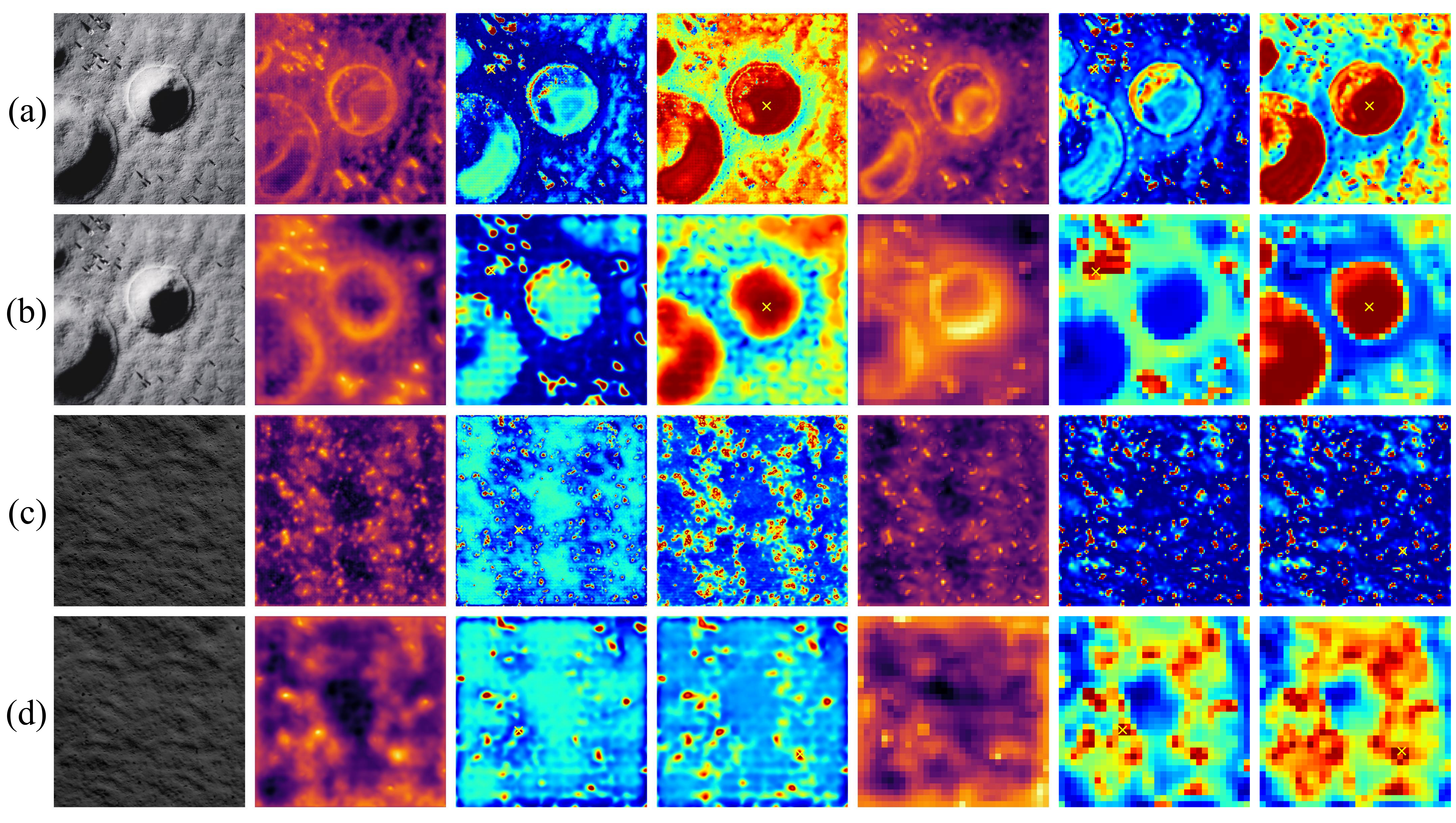}
    \caption{Emergent spatial awareness and semantic segmentation in the WARG framework. Qualitative visualization of internal representations on the LuSNAR (a, b) and South (c, d) datasets. Rows (a) and (c) utilize 1024 $\times$ 1024 input resolution, while (b) and (d) utilize 256 $\times$ 256 resolution. From left to right, the columns display: original satellite imagery; predicted saliency maps; and cosine similarity maps for a target point (marked by a yellow cross) at stride 1 and stride 8, respectively.}
    \label{fig:semantic_aware}
\end{figure*}

\begin{figure*}
    \centering
    \includegraphics[width=0.95\linewidth]{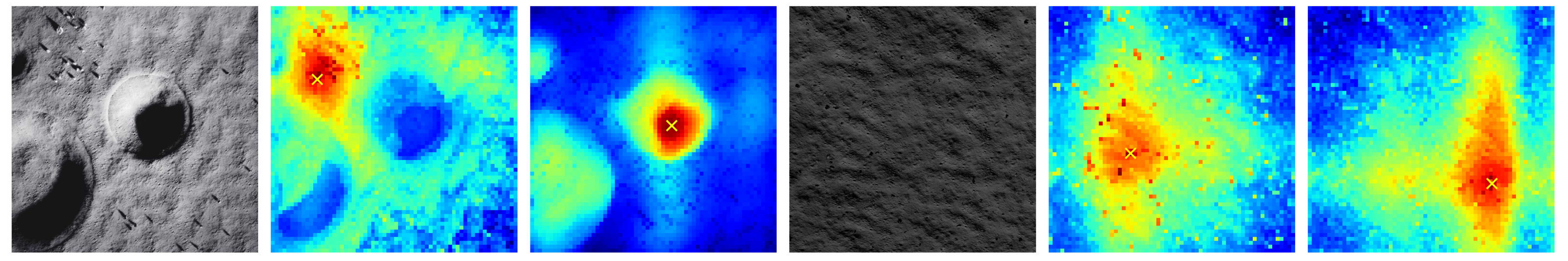}
    \caption{Feature identification limitations of the DINOv3 backbone. The first three panels display the original satellite imagery and corresponding cosine similarity maps for a target point (yellow cross) from the LuSNAR dataset, while the final three panels show the equivalent results for the South dataset.}
    \label{fig:dino_featmap}
\end{figure*}

To rigorously verify that WARG’s weight-sharing architecture successfully bridges the intra-entity divergence gap, we performed a quantitative sensitivity analysis on specific feature points (Fig.~\ref{fig:intra_entity_sim}). 
We manipulated two key environmental variables: illumination and spatial resolution.
The middle columns of Fig.~\ref{fig:intra_entity_sim} illustrate the model's response to extreme lighting perturbations. 
Even as the illumination coefficient $\alpha$ shifts toward complete darkness ($\alpha=0$) or extreme overexposure ($\alpha=1$), the cosine similarity between the corresponding rover and satellite features remains consistently high (typically $> 0.8$). 
This confirms that the shared encoder learns to suppress transient photometric variations in favor of shared stable descriptors.
The similarity curves in the rightmost columns of Fig.~\ref{fig:intra_entity_sim} demonstrate WARG's resilience to scale discrepancies, though some fluctuations are observed as resolution decreases toward the 256-pixel limit. 
This variance is primarily attributable to the loss of high-frequency details.
At lower resolutions, small-scale geographical features such as micro-rocks become sub-pixel or vanish entirely. 
Despite the disappearance of these fine-grained visual cues, the framework maintains a high baseline similarity, typically over 0.7.
These results quantitatively demonstrate that WARG projects heterogeneous cross-view inputs into a unified, viewpoint-invariant latent space, effectively solving the challenge of intra-entity divergence.

\begin{figure*}[!t]
    \centering
    \includegraphics[width=0.95\linewidth]{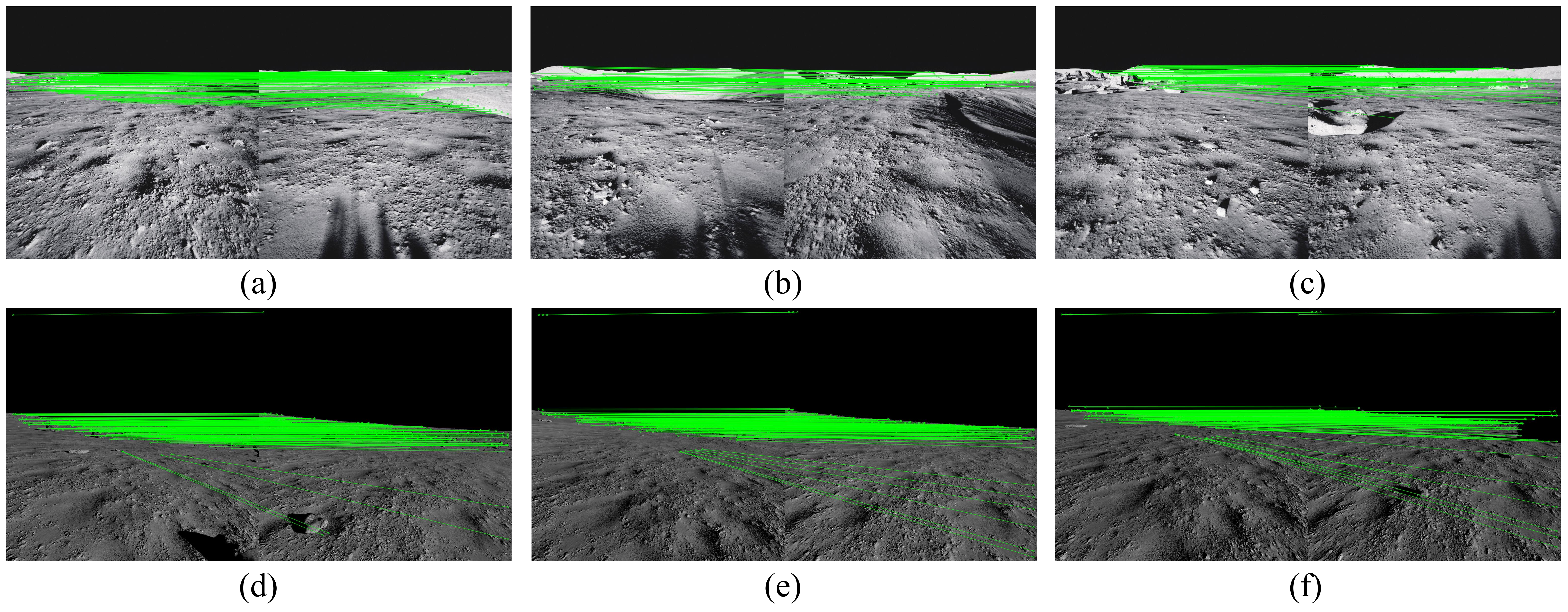}
    \caption{Accurate rover-view correspondence established through structural reasoning. Visualization of feature matching between temporally or spatially offset rover perspectives on the LuSNAR (a–c) and South (d–f) datasets.}
    \label{fig:structural_aware}
\end{figure*}

\subsection{Low-level Spatial Awareness}

We find that by learning from cross-view localization, WARG naturally develops low-level spatial awareness, including semantic and structural reasoning ability.

We first visualize the saliency map and feature similarity map predicted by WARG in Fig.~\ref{fig:semantic_aware}, compared to the one of backbone DINOv3 in Fig.~\ref{fig:dino_featmap}.
Although WARG is provided only with localization labels and no explicit segmentation supervision, the internal saliency maps (Fig.~\ref{fig:semantic_aware}, column 2) show that the model spontaneously learns to prioritize stable geomorphological landmarks, such as crater rims and rocks, over the uninformative background regolith.
This spatial awareness is further evidenced by feature similarity maps, particularly within the challenging South dataset imagery. 
In these scenarios, the model successfully isolates and delineates numerous small-scale rocks that are often obscured by extreme illumination, confirming the model's robust semantic reasoning ability.
This identification capability operates across a hierarchical spectrum through the model’s multi-scale feature architecture. 
While the high-resolution maps at stride 1 provide the sharp granularity necessary to segment small entities like individual rocks, the lower-resolution maps at stride 8 effectively aggregate broader contextual information to capture the global semantic extent of larger structures, such as expansive craters. 
By naturally balancing these multi-scale cues, WARG constructs a robust topographic prior that allows it to maintain sub-meter precision regardless of whether the available terrain landmarks are fine-grained boulders or large-scale geological formations.
In contrast, as illustrated in Fig.~\ref{fig:dino_featmap}, the DINOv3 backbone primarily identifies regions based on spatial proximity rather than semantic content, failing to achieve precise feature segmentation. 
This observation validates that the semantic awareness exhibited by WARG is a specialized capability developed through the cross-view localization task.

Apart from semantic reasoning, we notice that although WARG is trained exclusively for the cross-view localization task (ground-to-satellite), it develops a robust capability for same-view correspondence matching (ground-to-ground), as demonstrated in Fig.~\ref{fig:structural_aware}.
This cross-task generalization reveals that the model does not merely learn a mapping between two specific domains.
Instead, it captures intrinsic spatial structures that are fundamentally representative of the lunar landscape.
By forcing the model to align a horizontal ground perspective with a nadir orbital view during training, WARG is compelled to distill the spatial layout of the environment, a universal representation that remains valid regardless of the observation angle.
Consequently, the framework achieves highly accurate rover-view alignment through structural reasoning, effectively perceiving the arrangement of its surroundings even when the perspective shifts.
This emergent consistency within the rover's own frame of reference confirms that the model has developed a domain-agnostic spatial awareness, allowing it to maintain a coherent and reliable structural reasoning of its environment during active exploration.

\subsection{Ablation Study}

\begin{figure*}[!t]
    \centering
    \includegraphics[width=0.95\linewidth]{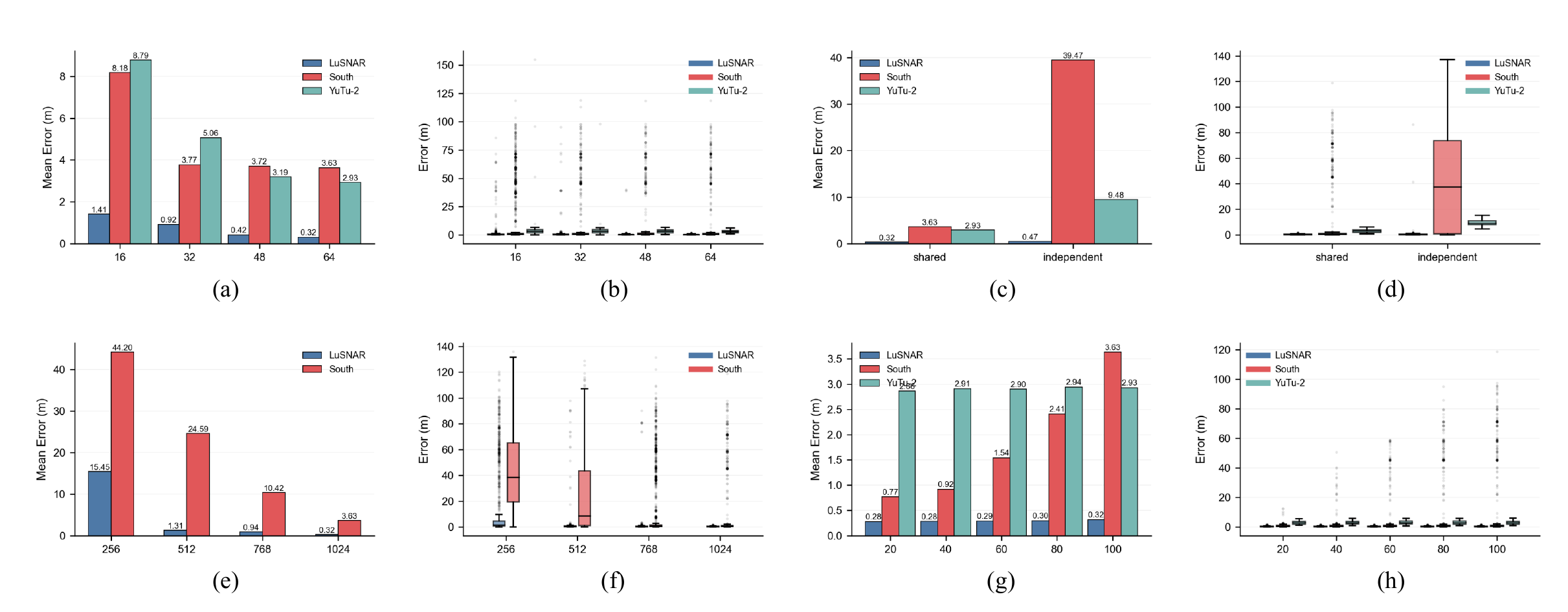}
    \caption{Comprehensive ablation studies quantifying architectural sensitivity and component contributions. (a)(b), Impact of graph node number, ranging from 16 to 64 nodes per feature map across four hierarchical strides (totaling 64 to 256 nodes). (c)(d) Comparison between weight-sharing and independent-weight architectures. (e)(f), Sensitivity to input resolution (256 to 1024 pixels). (g)(h), Influence of the prior search region size (20 m $\times $ 20 m to 100 m $\times$ 100 m).}
    \label{fig:ablation}
\end{figure*}

To verify the effectiveness of our architectural design, we conducted a systematic series of ablation experiments (Fig.~\ref{fig:ablation}).
First, we assessed the impact of structural density by varying the number of graph nodes per feature map (Fig.~\ref{fig:ablation}(a)(b)). 
We found that localization precision improves as the node density increases, reaching a plateau at 64 nodes per scale. 
Notably, the error distributions in Fig.~\ref{fig:ablation}(b) suggest that while the model maintains high baseline performance across all settings, increasing the node count significantly enhances robustness by suppressing outliers.
This confirms that a sufficiently dense structure provides the necessary structural redundancy to resolve inter-entity entanglement while remaining computationally efficient.
Second, a critical finding emerges from the comparison of the weight-sharing paradigm against an independent-weight baseline (Fig.~\ref{fig:ablation}(c)(d)). 
While independent weights achieve reasonable performance in standard terrains, they lead to a catastrophic failure on the South dataset, with mean errors surging to 39.47 m. 
This collapse confirms that the shared-weight architecture acts as a vital inductive bias, enforcing the extraction of viewpoint-invariant features that are essential for bridging the extreme domain gap inherent in lunar environments.
Third, the resolution analysis (Fig.~\ref{fig:ablation}(e)(f)) confirms that while precision scales with input size, WARG maintains operational stability even at 512 pixels as the median error remains consistently low, offering flexibility for different hardware constraints.
Finally, we evaluated the model’s robustness to spatial uncertainty by expanding the prior search region from 20 m $\times$ 20 m to 100 m $\times$ 100 m (Fig.~\ref{fig:ablation}(g)(h)). 
The marginal increase in error across this five-fold expansion of the search space demonstrates that WARG’s structural matching ability is remarkably resilient to initial positioning errors, providing a robust solution for practical autonomous exploration.

\subsection{Robustness Study}

\begin{figure*}[!t]
    \centering
    \includegraphics[width=0.95\linewidth]{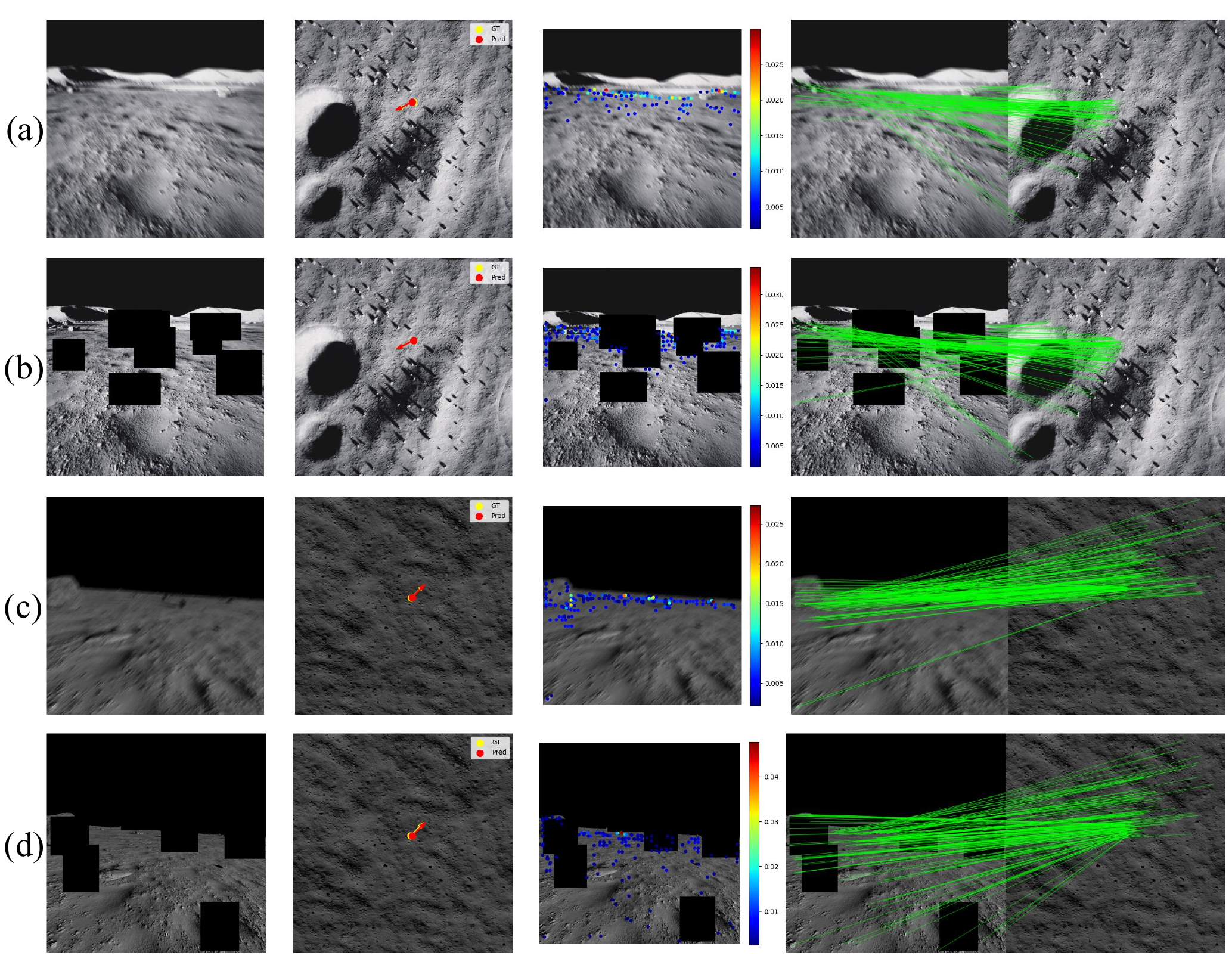}
    \caption{Resilience of WARG to perceptual degradation via motion blur and feature occlusion. Qualitative localization performance is evaluated on the LuSNAR (a, b) and South (c, d) datasets under adversarial conditions. (a)(c), Robustness to motion blur. (b)(d), Robustness to partial occlusion.}
    \label{fig:robustness}
\end{figure*}

To assess the operational reliability of WARG in real-world exploration scenarios, we stress-tested the framework against significant perceptual degradations, specifically simulated motion blur and stochastic feature occlusion (Fig.~\ref{fig:robustness}). 
In the presence of severe motion blur (Fig.~\ref{fig:robustness}(a)(c)), which typically causes conventional feature-matching algorithms to fail due to the smearing of high-frequency textures, WARG continues to achieve highly accurate localization. 
This resilience suggests that the framework has learned to extract a persistent structure that remains robust to blur, rather than relying solely on high-frequency details.
Furthermore, the model demonstrates sophisticated fault tolerance under partial occlusion (Fig.~\ref{fig:robustness}(b)(d)). 
By deliberately masking prominent geomorphological entities like craters and rocks, we simulate challenging conditions such as lens obstruction, sensor failure, or dust interference. 
Despite these missing inputs, WARG's structural reasoning enables it to infer the rover’s position by leveraging the structure formed by the surviving landmarks. 
The near-perfect alignment between predicted and ground-truth markers under these adversarial conditions confirms that WARG does not rely on any single local landmark.
Instead, it develops a robust, holistic understanding of the environmental layout, ensuring navigation safety even when the visual data is incomplete or corrupted.

\section{Discussion}

\subsection{Interpretation of Results}
Our method achieves a significant reduction in localization error compared to current SOTA approaches across both simulated and real-world datasets.
This performance gain is primarily driven by our leveraging of graph-based representations.
By integrating deep-learning-based semantic embeddings with geometry-driven structural representations, the model effectively captures both appearance and spatial layout cues. 
The two graph learning stages of our method, graph construction and graph alignment, contribute synergistically to this improvement.

\subsection{Practical Application Potential}
The demonstrated accuracy, efficiency, and robustness suggest that our approach has significant practical implications for lunar and planetary exploration. 
With a reduced mean localization error close to one pixel and high processing efficiency, our method can support long-term navigation of rovers in GNSS-denied and feature-sparse environments. 
Its resilience to extreme illumination shifts, scale discrepancies, motion blur, and partial occlusions further enhances its suitability for real-world deployment, enabling global-scale exploration and mapping tasks that are otherwise infeasible using traditional local localization methods.

\subsection{Emergent Spatial Awareness}
Beyond localization, we observe that the model naturally develops low-level spatial awareness, including semantic and structural reasoning, simply by learning the cross-view localization task. 
This suggests that cross-view localization could serve as a foundational pretext task, analogous to next-token prediction in natural language processing~\citep{sequence}, for acquiring broader spatial intelligence. 
The model’s ability to infer semantic categories, understand spatial arrangements, and establish correspondences across arbitrary viewpoints demonstrates its potential to generalize to diverse spatial reasoning tasks with minimal additional supervision.

\begin{figure*}[!t]
    \centering
    \includegraphics[width=0.95\linewidth]{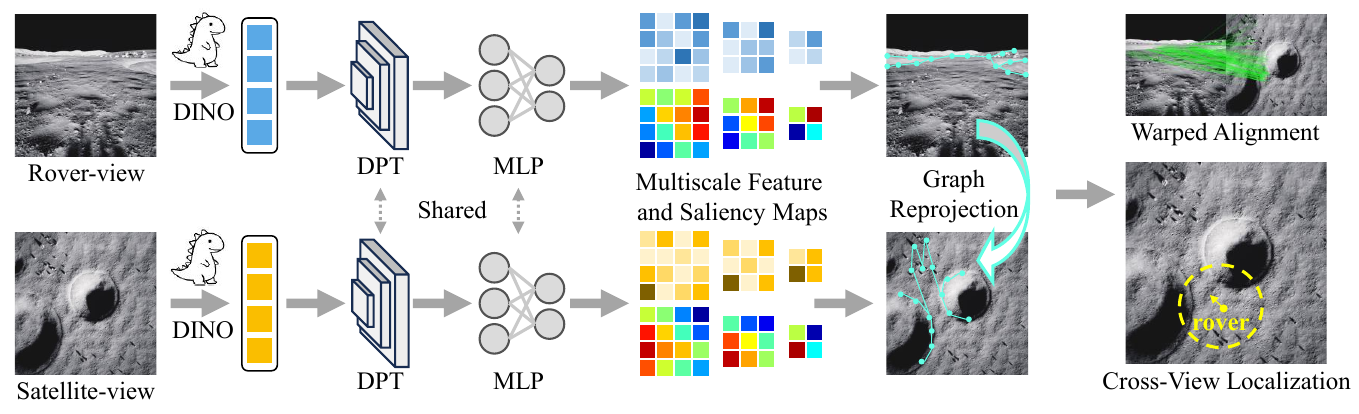}
    \caption{Overall framework of WARG, which consists of three main stages: feature extraction, graph construction, and graph alignment.}
    \label{fig:framework}
\end{figure*}

\subsection{Limitations and Future Research}
Despite these advancements, our method still has certain limitations. 
First, localization accuracy degrades in scenarios where no informative terrain overlaps exist between the rover and satellite imagery. 
This limitation is intrinsic to cross-view localization. 
Future work could incorporate complementary information or strategies from local localization methods to handle such outlier cases, while maintaining the high accuracy of global localization.
Second, while the model exhibits emergent spatial awareness, this capability is currently constrained. 
Semantic segmentation is limited to a few classes due to the sparse and repetitive lunar terrain.
Expanding the training regime to include diverse terrestrial and Martian datasets could enhance the semantic diversity and the model's spatial understanding ability.
Finally, we aim to explore the integration of view-synthesis frameworks~\citep{deepstereo, nerf} to transition from a discriminative paradigm to a generative one, enabling the model to predict scene layouts directly from spatial reasoning.

\section{Method}

\subsection{Problem Formulation}

We address the problem of lunar cross-view localization by cross-referencing horizontal rover-view observations with orbital satellite imagery. 
Let the geometric configuration of a viewing agent be denoted by the tuple $\mathcal{C} = \{\mathbf{K}, \mathbf{R}, \mathbf{t}\}$, comprising the intrinsic matrix $\mathbf{K} \in \mathbb{R}^{3 \times 3}$, the rotation matrix $\mathbf{R} \in SO(3)$, and the translation vector $\mathbf{t} \in \mathbb{R}^3$ defined in a unified selenocentric world coordinate system.

The input consists of a rover query $\mathcal{Q} = (\mathbf{I}_r, \mathbf{D}_r, \mathcal{C}_r)$ and a satellite reference $\mathcal{S} = (\mathbf{I}_s, \mathcal{C}_s)$. Here, $\mathbf{I}_r$ and $\mathbf{I}_s$ denote the rover and satellite images, respectively, and $\mathbf{D}_r$ represents the rover-view depth map estimated via stereo vision or LiDAR.
The rover's rotation $\mathbf{R}_r$ (rover-to-world) and intrinsic parameters $\mathbf{K}_r$ are assumed known, while the initial translation guess is denoted as $\mathbf{t}_r^{\text{init}}$. 
The satellite parameters $\mathcal{C}_s$ are treated as known constants given the high precision of orbital data.

Our objective is to estimate the optimal rover translation $\mathbf{t}_r$ based on the initial guess $\mathbf{t}_r^{\text{init}}$ that maximizes the semantic and structural consistency between the cross-view pair.

\subsection{Overall Framework}

Fig.~\ref{fig:framework} illustrates the overall framework of WARG, which comprises three stages: feature extraction, graph construction, and graph alignment.
In the feature extraction stage, we utilize DINOv3 with DPT~\citep{dpt} to extract multiscale feature maps.
Subsequently, in the graph construction stage, WARG captures salient and informative terrain features and constructs a structured graph from the rover-view image.
Finally, the graph alignment stage establishes accurate correspondences by reprojecting the rover-view graph into the satellite domain and thereby enforcing structural constraints to predict the optimal rover position.
The following sections introduce these three stages in detail.

\subsection{Feature Extraction}
To obtain robust and semantically rich representations of the lunar terrain, we employ a vision transformer-based encoder-decoder architecture. 
The model receives a rover-view image $\mathbf{I}_r \in \mathbb{R}^{H \times W \times 3}$ and a satellite-view image $\mathbf{I}_s \in \mathbb{R}^{H \times W \times 3}$ as inputs. 

We utilize DINOv3 as the backbone encoder to extract high-level semantic tokens, leveraging its strong generalization capabilities in unstructured environments. 
To recover spatial details lost during the tokenization process, we attach a DPT head. 
The DPT module reassembles the transformer tokens into image-like feature maps at varying resolutions, enabling the framework to capture both holistic terrain context and fine-grained texture details.

Formally, for a given input image $\mathbf{I}$, the feature extraction module outputs a multi-scale feature pyramid, denoted as:

\begin{equation}
    \mathcal{F} = \left\{ \mathbf{F}^{(s)} \mid s \in \mathcal{S}=\{1, 2, 4, 8\} \right\},
\end{equation}

\noindent where $s$ represents the stride relative to the input resolution. 
Consequently, each feature map $\mathbf{F}^{(s)}$ possesses spatial dimensions of $\frac{H}{s} \times \frac{W}{s}$ with a channel dimension $C$. 
This hierarchical representation facilitates a coarse-to-fine alignment strategy: coarse features ($\mathbf{F}^{(8)}$) are utilized to register macroscopic landmarks such as impact craters, while high-resolution features ($\mathbf{F}^{(1)}$) resolve fine-grained details like individual rocks to ensure precise correspondence.
Crucially, the DINOv3 backbone remains frozen, whereas the DPT head is trainable. 
By sharing the DPT weights across both viewpoints, we enforce a shared feature space that promotes the learning of viewpoint-invariant representations.

\subsection{Graph Construction}
\label{subsec:graph_fusion}

To distill the dense feature pyramid into a structured geometric representation, we introduce a graph construction module that identifies and encodes salient terrain entities. 
This process relies on two lightweight Multi-Layer Perceptron (MLP) heads: a saliency head ($\Psi_{\text{sal}}$) and a feature refinement head ($\Psi_{\text{feat}}$). 
Both heads utilize GELU activation functions and, crucially, share weights across all scales and viewpoints. 
This architectural constraint enforces scale-invariant feature learning, ensuring that the model maintains robust representations across varying resolutions without overfitting to a specific scale and perspective.

\subsubsection{Saliency and Feature Refinement}
Given a feature map $\mathbf{F}^{(s)}$ at scale $s$, the heads operate pixel-wise to generate a saliency weight map $\mathbf{M}^{(s)}$ and a refined feature descriptor map $\mathbf{D}^{(s)}$:

\begin{equation}
    \mathbf{M}^{(s)} = \Psi_{\text{sal}}(\mathbf{F}^{(s)}), \quad 
    \mathbf{D}^{(s)} = \Psi_{\text{feat}}(\mathbf{F}^{(s)}),
\end{equation}

\noindent where the higher value in the saliency weight map $\mathbf{M}^{(s)}$ indicates greater visual saliency, corresponding to features with higher discriminative power for the localization task.

\subsubsection{Rover-View Graph Construction}
Given the significant geometric asymmetry between the two viewpoints, where the satellite orbital imagery encompasses a vast spatial extent compared to the localized rover field of view, we first construct a sparse graph $\mathcal{G}_r = (\mathcal{V}_r, \mathcal{E}_r)$ from rover view to minimize computational redundancy while retaining critical navigation cues. 
We employ an importance sampling strategy based on the generated saliency maps. For each scale $s \in \mathcal{S}$, we select the top-$N/|\mathcal{S}|$ pixels with the highest saliency scores in $\mathbf{M}_r^{(s)}$.

Each selected pixel $i$ constitutes a node $v_i \in \mathcal{V}_r$, characterized by the tuple:
\begin{equation}
    v_i = \left( \mathbf{u}_i, \mathbf{f}_i, w_i \right),
\end{equation}
\noindent where $\mathbf{u}_i$ represents the pixel coordinate referenced to the original image frame, $\mathbf{f}_i \in \mathbb{R}^C$ is the refined feature vector sampled from $\mathbf{D}_r^{(s)}$, and $w_i$ is the associated saliency weight from $\mathbf{M}_r^{(s)}$:
\begin{equation}
    \mathbf{f}_i = \mathbf{D}_r^{(s)}(\mathbf{u}_i), \quad w_i = \mathbf{M}_r^{(s)}(\mathbf{u}_i).
\end{equation}
Aggregating across all four scales results in a multi-resolution node set with $|\mathcal{V}_r| = N$.

We define the graph edges $\mathcal{E}_r$ based on the 3D physical layout of the lunar surface. 
For each node $v_i$, we project its 2D pixel coordinate $\mathbf{u}_i$ to a 3D point $\mathbf{P}_i$ in the rover-centric coordinate frame using the provided depth map $\mathbf{D}_r$ and camera intrinsics $\mathbf{K}_r$:

\begin{equation}
    \mathbf{P}_i = \mathbf{D}_r(\mathbf{u}_i) \cdot \mathbf{K}_r^{-1} \tilde{\mathbf{u}}_i,
\end{equation}

\noindent where $\tilde{\mathbf{u}}_i$ is the homogeneous representation of $\mathbf{u}_i$. The edge $\mathbf{e}_{ij} \in \mathcal{E}_r$ between nodes $v_i$ and $v_j$ is defined as the relative 3D displacement vector:

\begin{equation}
    \mathbf{e}_{ij} = \mathbf{P}_j - \mathbf{P}_i.
\end{equation}

This formulation ensures that the graph captures the physical structure of the terrain.

\subsubsection{Satellite-View Dense Representation}
For the satellite view, we apply the same shared heads ($\Psi_{\text{sal}}, \Psi_{\text{feat}}$) to generate multiscale saliency and feature maps. 
However, unlike the rover view, we retain the dense representation without sampling. This preserves the global context required to ground the sparse rover observations against the orbital map during the subsequent alignment stage.

\subsection{Graph Alignment}
The graph alignment stage evaluates a set of translation hypotheses by measuring the structural and semantic consistency between the sparse rover graph and the dense satellite feature pyramid. 

\subsubsection{Graph Reprojection}
Given an initial rover translation guess $\mathbf{t}_r^{\text{init}}$, we define a spatial search region and sample $K$ translation candidates $\mathbf{T} = \{\mathbf{t}_k\}_{k=1}^K$. For each candidate $\mathbf{t}_k$, we reproject the 3D rover-frame coordinates $\mathbf{P}_i$ of all $N$ graph nodes into the satellite image plane. The reprojected coordinate $\mathbf{u}_{i,k}$ for node $i$ under candidate $\mathbf{t}_k$ is given by:
\begin{equation}
    \mathbf{u}_{i,k} = \mathbf{K}_s \mathbf{R}_s^\top \left[ (\mathbf{R}_r \mathbf{P}_i + \mathbf{t}_k) - \mathbf{t}_s \right].
\end{equation}

\subsubsection{Multi-Scale Feature Sampling}
For each candidate $\mathbf{t}_k$, we sample the satellite feature maps $\mathbf{F}_s^{(s)}$ and saliency maps $\mathbf{M}_s^{(s)}$ at the projected locations $\mathbf{u}_{i,k}$ across all scales $s \in \mathcal{S}$:
\begin{equation}
    \hat{\mathbf{f}}_{i,k}^{(s)} = \mathbf{F}_s^{(s)}(\mathbf{u}_{i,k}), \quad \hat{w}_{i,k}^{(s)} = \mathbf{M}_s^{(s)}(\mathbf{u}_{i,k}).
\end{equation}
This multi-scale sampling strategy ensures scale-invariant awareness, allowing the framework to simultaneously resolve macroscopic topographical structures, such as impact craters, and fine-grained primitives, such as individual boulders.

\subsubsection{Cross-Scale Joint Saliency Alignment}
To prioritize landmarks that are mutually salient across both viewpoints, we define a joint reliability weight $\beta_{i,k}^{(s)}$ for each node $i$ at scale $s$:
\begin{equation}
    \beta_{i,k}^{(s)} = w_i + \hat{w}_{i,k}^{(s)},
\end{equation}
where $w_i$ is the intrinsic saliency of the rover node. Crucially, we normalize these weights across the entire multi-scale node set using a global softmax:
\begin{equation}
    \alpha_{i,k}^{(s)} = \frac{\exp(\beta_{i,k}^{(s)})}{\sum_{s' \in \mathcal{S}} \sum_{j=1}^{N} \exp(\beta_{j,k}^{(s')})}.
\end{equation}
This normalization allows the model to dynamically reweight the contribution of different terrain features and scales, effectively focusing on the most informative correspondences within the $N \times |\mathcal{S}|$ total observations.

The alignment logit for a candidate $\mathbf{t}_k$ is then computed as the weighted sum of cosine similarities:
\begin{equation}
    \mathcal{L}_k = \sum_{s \in \mathcal{S}} \sum_{i=1}^N \alpha_{i,k}^{(s)} \cdot \left(\frac{\mathbf{f}_i^\top \hat{\mathbf{f}}_{i,k}^{(s)}}{\|\mathbf{f}_i\| \| \hat{\mathbf{f}}_{i,k}^{(s)} \| }  \right).
\end{equation}

\subsubsection{Inference and Training}
The final probability distribution over the translation candidates is obtained by applying a temperature-scaled softmax:
\begin{equation}
    P(\mathbf{t}_k | \mathbf{I}_r, \mathbf{I}_s) = \frac{\exp(\mathcal{L}_k \cdot e^{\tau})}{\sum_{k'=1}^K \exp(\mathcal{L}_{k'} \cdot e^{\tau})},
\end{equation}
where $\tau$ is a learnable log-temperature parameter.

For inference, the model outputs the candidate with the maximum probability: $\mathbf{t}_r^* = \arg\max_{\mathbf{t}_k} P(\mathbf{t}_k)$. 

For training, we identify the candidate $\mathbf{t}^+$ that is spatially closest to the ground-truth translation $\mathbf{t}_{GT}$ and optimize the network using the Negative Log-Likelihood (NLL) loss:
\begin{equation}
    \mathcal{L}_{\text{NLL}} = -\log \frac{\exp(\mathcal{L}_+)}{\sum_{k=1}^K \exp(\mathcal{L}_k)}.
\end{equation}
This objective function constrains the model to maximize the similarity of the true rover position by enforcing structural and semantic consistency between the cross-view observations.

\section{Related Works}
Lunar rover localization is fundamental for autonomous navigation and scientific exploration on the Moon.
In the absence of GNSS, localization relies on onboard sensors, with most prior missions adopting local approaches that estimate motion with respect to previous positions~\citep{chang-e-6-1,artemis-1,chandrayaan}.
Typical methods such as inertial navigation and wheel odometry use accelerometer, gyroscope, and encoder data but are highly susceptible to noise, bias drift, and wheel slippage, leading to cumulative position errors~\citep{inertial,yutu-2,chandrayaan-3}.
Stereo-based localization improves short-term robustness by reconstructing sparse 3D points and estimating motion through feature or point cloud matching between consecutive frames~\citep{stereo-1,stereo-2}.
However, all these local methods inevitably suffer from long-term drift, where small estimation errors accumulate into kilometer-scale deviations, severely limiting their effectiveness for large-scale lunar exploration~\citep{stereo-drift}.

To overcome this limitation, recent studies have increasingly focused on cross-view localization, a promising approach capable of achieving global positioning similar to GNSS~\citep{vigor,cvl}, particularly in terrestrial settings.
These methods estimate the ground camera's absolute position in a global coordinate frame by matching ground-view and satellite-view imagery, thus removing dependence on prior motion estimation and mitigating drift accumulation.
A common strategy is cross-view retrieval, where the ground-view image is used to retrieve the most similar satellite-view image from a database, with the corresponding satellite coordinate serving as the localization result~\citep{cross-transformer,co-retrieval,sample4geo}.
However, retrieval-based methods provide only coarse localization, as their accuracy is constrained by the spatial resolution and granularity of available satellite imagery.
To achieve higher precision, recent works aim for fine-grained localization, such as predicting pixel-wise correspondences or directly estimating the rover's pose~\citep{beyond}.
Most methods follow a common paradigm: converting the ground imagery into a BEV map and then performing BEV-to-satellite matching.
The BEV map is typically generated either by lifting image features along the depth axis and splatting them onto a ground plane~\citep{lift,orienternet,snap,lcv-1}, or by lifting a BEV grid in height and projecting it back into the ground-view domain~\citep{bevformer,fg2,uncertainty}.
However, these dense 2D or 3D constructions are computationally inefficient and redundant in the lunar scenario, given the inherently sparse distribution of informative features in lunar terrain.
Furthermore, the asymmetric and independent processing of the two viewpoints exacerbates intra-entity divergence, as models tend to prioritize viewpoint-specific details over invariant features.
Moreover, many existing matching strategies rely exclusively on point-to-point correspondence, neglecting the structural relationships among features, which are essential for mitigating inter-entity entanglement and achieving robust and accurate localization in challenging lunar environments~\citep{fg2,revisit}.

Despite the existing challenges in cross-view localization, it plays a crucial role in enabling future large-scale lunar exploration missions, such as China’s Chang’e-7~\citep{chang-e-7} and other multi-robot or long-duration surface operations.
By providing global and drift-free positioning without relying on external infrastructure like GNSS, it remains essential for realizing next-generation lunar missions that aim to explore permanently shadowed regions, conduct in-situ resource utilization, and establish sustainable lunar bases~\citep{apollo}.

\bibliography{main}

\end{document}